\documentclass[letterpaper, 10 pt, conference]{ieeeconf}  

\IEEEoverridecommandlockouts                              

\usepackage{graphics} 
\usepackage{epsfig} 
\usepackage{times} 
\usepackage{amsmath} 
\usepackage{amssymb}  
\let\proof\relax 
\let\endproof\relax
\usepackage{amsthm}
\usepackage[T1]{fontenc}
\usepackage{soul}
\usepackage{url}
\usepackage[hidelinks]{hyperref}
\usepackage[utf8]{inputenc}
\usepackage[small]{caption}
\usepackage{booktabs}
\usepackage{algorithm}
\usepackage[noend]{algpseudocode}
\usepackage{algorithmicx}
\usepackage[switch]{lineno}
\usepackage{subcaption}
\usepackage{cleveref}

\usepackage[dvipsnames]{xcolor}

\algnewcommand\algorithmicinput{\textbf{Input:}}
\algnewcommand\INPUT{\item[\algorithmicinput]}
\algnewcommand\algorithmicoutput{\textbf{Output:}}
\algnewcommand\OUTPUT{\item[\algorithmicoutput]}

\title{\LARGE \bf
Caching-Augmented Lifelong Multi-Agent Path Finding
}

\author{Yimin Tang$^{1*}$, Zhenghong Yu$^{2*}$, Yi Zheng$^1$, T. K. Satish Kumar$^1$, Jiaoyang Li$^3$, Sven Koenig$^1$
\thanks{$^{*}$Equal Contribution}%
\thanks{$^{1}$University of Southern California, 3650 McClintock Ave, Los Angeles, CA 90089
        {\tt\small \{yimintan,yzheng63\}@usc.edu, tkskwork@gmail.com, skoenig@usc.edu}}%
\thanks{$^{2}$ShanghaiTech University, 393 Middle Huaxia Road, Shanghai 201210
        {\tt\small yuzhh1@shanghaitech.edu.cn}}%
\thanks{$^{3}$Carnegie Mellon University, 5000 Forbes Ave, Pittsburgh, PA 15213
        {\tt\small jiaoyanl@andrew.cmu.edu}}%
}%

\begin{document}

\maketitle
\thispagestyle{empty}
\pagestyle{empty}

\begin{abstract}

Multi-Agent Path Finding (MAPF), which involves finding collision-free paths for multiple robots, is crucial in various applications. Lifelong MAPF, where targets are reassigned to agents as soon as they complete their initial targets, offers a more accurate approximation of real-world warehouse planning.
In this paper, we present a novel mechanism named Caching-Augmented Lifelong MAPF (CAL-MAPF), designed to improve the performance of Lifelong MAPF. We have developed a new type of map grid called cache for temporary item storage and replacement, and created a locking mechanism to improve the planning solution's stability. A task assigner (TA) is designed for CAL-MAPF to allocate target locations to agents and control agent status in different situations. CAL-MAPF has been evaluated using various cache replacement policies and input task distributions. We have identified three main factors significantly impacting CAL-MAPF performance through experimentation: suitable input task distribution, high cache hit rate, and smooth traffic. In general, CAL-MAPF has demonstrated potential for performance improvements in certain task distributions, map and agent configurations.

\end{abstract}

\section{INTRODUCTION}

Automated warehouses represent a multibillion-dollar industry led by corporations such as Amazon, Ocado, and inVia. These facilities deploy hundreds of robots to transport goods from one location to another~\cite{wurman2008coordinating}. Planning collision-free solutions for hundreds of robots is a key component of these automated warehouses, and can be simplified to a multi-agent path finding problem (MAPF)~\cite{stern2019multi}. The MAPF problem requires planning collision-free paths for multiple agents from their start locations to pre-assigned target locations in a known environment while minimizing a given cost function. Many algorithms have been developed to solve this problem optimally and suboptimally, such as Conflict Based Search (CBS)~\cite{sharon2015conflict}, \(M^*\),~\cite{wagner2011m} and Enhanced CBS (ECBS)~\cite{barer2014suboptimal}. 

Although MAPF is classified as an NP-hard problem~\cite{yu2013structure}, it still represents a simplistic approximation of actual warehouse planning. It is the 'one-shot' version of the real application challenge, where an agent only needs to reach one target location and remains stationary until every agent has arrived at their respective targets. To address this limitation, a new problem called Multi-Agent Pickup and Delivery (MAPD) or Lifelong MAPF~\cite{lifelong2017} has been introduced. This version assigns new targets to agents once they reach their current targets, reflecting continuous operational scenarios better. Various algorithms, such as RHCR~\cite{li2021lifelong}, MAPF-LNS~\cite{li2021anytime}, PIBT~\cite{okumura2022priority}, and LaCAM~\cite{okumura2023lacam}, have been created and applied in practical environments.

Although many studies focus on developing effective algorithms, there is also interest in improving the throughput of automated warehouses within the Lifelong MAPF framework by optimizing warehouse layouts. Prominent approaches include Fishbone design~\cite{fishbone2009}, DSAGE~\cite{ijcai2023p611}, and GGO~\cite{zhang2024guidance}. 
However, these works mainly focus on static storage design and do not consider input distributions with specific patterns that may change over time. To address this problem, some studies have explored intelligent decision-making strategies~\cite{tang2019strategic,yuan2019velocity,li2020storage}. Notable examples include Kiva system and various policies, such as the velocity-based policy~\cite{yuan2019velocity}, and mixed-integer optimization and heuristic methods~\cite{WeidingerBB18}. However, these methods are designed for specific warehouse systems, making applying them to different warehouses difficult. They also require implementing complex rearrangement policies for the entire storage.

In this paper, we present a dynamic cache delivery mechanism named Caching-Augmented Lifelong MAPF (CAL-MAPF), inspired by cache design, a foundational idea extensively utilized in computer architecture, databases, and various other domains. This cache mechanism is adaptable to multiple Lifelong MAPF algorithms, warehouse storage strategies, and incoming task distributions. We developed a new map grid type that allows items to be temporarily stored and replaced. We also designed a lock mechanism to enhance the stability of the planning solution and a task assigner (TA), which allocates target locations to agents and controls agent status in different situations. We have evaluated this cache mechanism using different cache replacement policies and input task distributions. Through experimentation, we identified three main factors that significantly impact CAL-MAPF performance: input task distribution, cache hit rate, and map design. In general, CAL-MAPF has demonstrated potential for performance improvements in certain task distributions, map and agent configurations.

\section{Problem Definition}

Lifelong Multi-Agent Path Finding (Lifelong MAPF) problem presents unique challenges due to the dynamic and unending nature of the tasks.  
Let $I=\{1,2,\cdots,N\}$ denote a set of $N$ agents. $G = (V,E)$ represents an undirected graph, where each vertex $v \in V$ represents a possible location of an agent in the workspace, and each edge $e \in E$ is a unit-length edge between two vertices that moves an agent from one vertex to the other. Self-loop edges are allowed, which represent ``wait-in-place'' actions. Each agent $i\in I$ has a unique start location $s_i \in V$. Assume there is an task queue \(Q = [q_1, q_2, ...]\). For each task \(q_i \in Q\), \(q_i = (idx, loc)\) represents a certain kind of item with type number $idx$ should be sent to position $loc$. This $idx$ item should be located in a reachable vertex in \(V\) where at least one agent can find a path to the vertex. Each item must be delivered to a vertex in \(V\) as the user specifies. These tasks are generated from a specific probability distribution to enhance realism.

\begin{figure}[t!]
\centering
\includegraphics[width=0.48\textwidth]{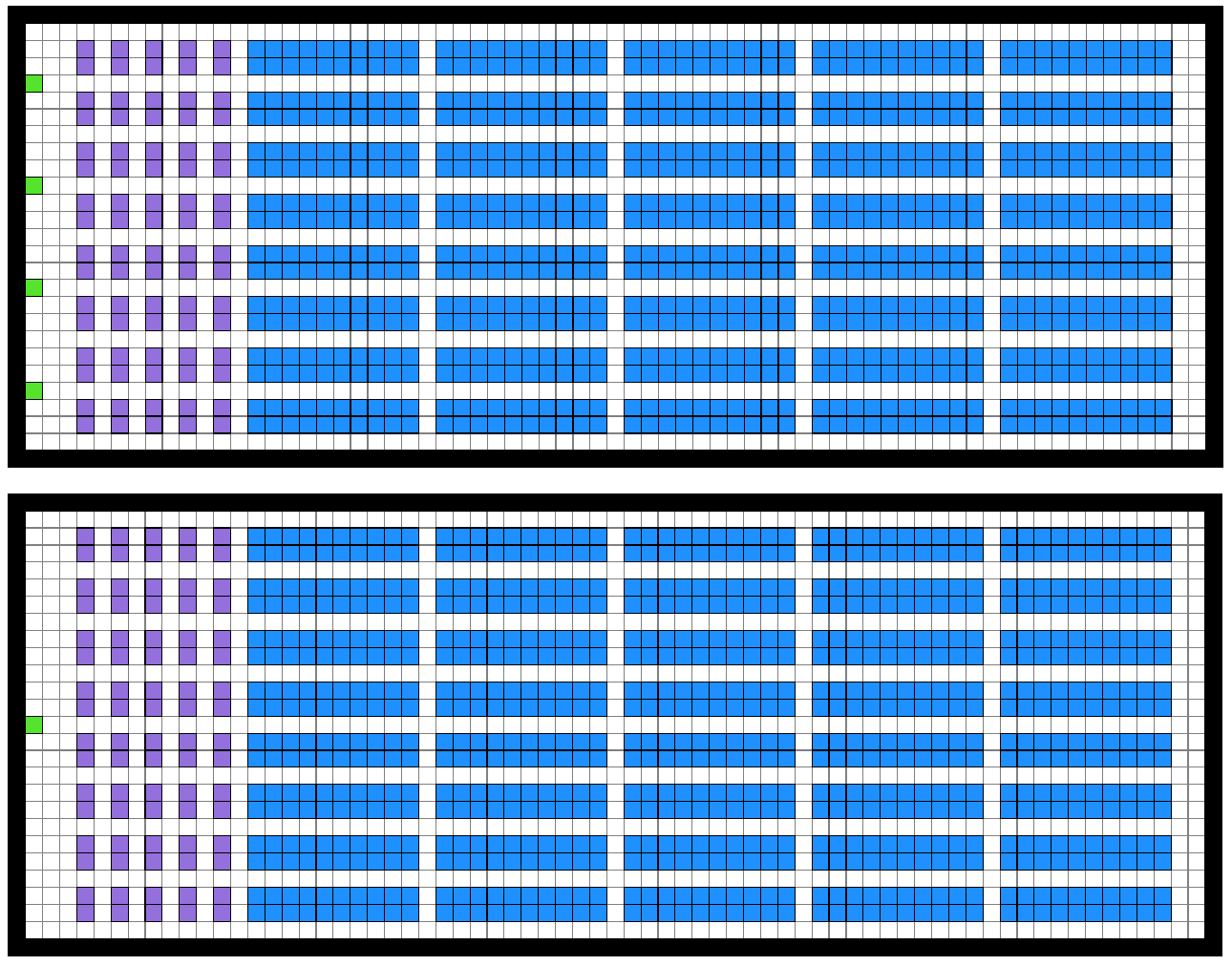} 
\caption{Caching-Augmented Maps: (1) Blue grids represent Shelves \(S\). (2) Purple grids represent Caches \(C\). (3) Green grids represent unloading ports \(U\). The upper map has multiple ports, while a single port is in the bottom one. In the multi-port map, each unloading port has an independent cache area, task queue, and agents. The cache areas are near the unloading ports within \(\pm 2\) rows. For the single-port map, the port can utilize all agents and all caches. Given that the number of cache grids can affect the cache hit rate, we also tested different numbers of cache grids, ranging from 80 to 16, by removing cache grids column by column from right to left.
}
\label{fig:cal_mapf}
\end{figure}

We study an online setting in which incoming tasks are not known in advance. It is assumed that there is a task assigner (external to MAPF algorithm). The TA determines the specific target locations for agents based on tasks in \(Q\). In our paper, TA is considered naive and will return the target location based on the first unassigned task in \(Q\) for each spare agent. 
Each agent \( i \in I \) starts from a location \( s_i \in V \) and its target location depends on the TA. Each action of agents, which could be waiting in place or moving to an adjacent vertex, takes a unit of time. An agent's path, \( p^i = [v_0^i, v_1^i, \ldots, v_{T_i}^i] \), tracks the sequence of vertices traversed by agent \( i \) from its start to its target. As agents complete their paths at their assigned targets, they are immediately assigned new target locations, reflecting warehouse operations' perpetual and iterative nature. 

As shown in ~\Cref{fig:cal_mapf}, we primarily focus on 2D warehouse layout maps and categorize all non-obstacle map grids into three types: blue grids \(B=\{b_i | b_i \in V\}\) for shelves that store items, white grids \(W=\{w_i | w_i \in V\}\) for normal aisle grids, and green grids \(U=\{u_i | u_i \in V\}\) for unloading ports where agents deliver items to complete a task. Agents can enter \(B\) positions only from \(W\) positions. Direct transitions between two positions in \(B\) are not allowed.

\section{Related work}

\subsection{MAPF}

(One-Shot) MAPF, which has been proved an NP-hard problem with optimality~\cite{yu2013structure}, has a long history~\cite{silver2005cooperative}. This problem is finding collision-free paths for multiple agents while minimizing a given cost function. It has inspired a wide range of solutions for its related challenges. 
Decoupled strategies, as outlined in \cite{silver2005cooperative,luna2011push,wang2008fast}, approach the problem by independently planning paths for each agent before integrating these paths. 
In contrast, coupled approaches \cite{standley2010finding,standley2011complete} devise a unified plan for all agents simultaneously. There also exist dynamically coupled methods~\cite{sharon2015conflict,wagner2015subdimensional} that consider agents planned independently at first and then together only when needed in order to resolve agent-agent conflicts. 
Among these, Conflict-Based Search (CBS) algorithm \cite{sharon2015conflict} stands out as a centralized and optimal method for MAPF, with several bounded-suboptimal variants such as ECBS~\cite{barer2014suboptimal} and EECBS~\cite{li2021eecbs}. Some suboptimal MAPF solvers, such as Prioritized Planning (PP)~\cite{erdmann1987multiple,silver2005cooperative}, PBS~\cite{ma2019searching}, PIBT~\cite{okumura2022priority} and their variant methods~\cite{chan2023greedy,li2022mapf,okumura2023lacam,okumura2023lacam3} exhibit better scalability and efficiency.

\subsection{Lifelong MAPF}

Compared to the MAPF problem, Lifelong MAPF continuously assigns new target locations to agents once they have reached their current targets. Agents don't need to arrive at their targets simultaneously in Lifelong MAPF. There are mainly three kinds of methods: solving the problem as a whole~\cite{nguyen2019generalized}, using MAPF methods but replanning all paths at each specified length timestep~\cite{li2021lifelong,okumura2022priority}, and replanning only when agents reach their current target locations and are assigned new targets~\cite{vcap2015complete,grenouilleau2019multi,okumura2023lacam,okumura2023lacam3}. Among them, MAPF-LNS~\cite{li2021anytime,li2022mapf} and LaCAM~\cite{okumura2023lacam,okumura2023lacam3} are the leading methods.

\subsection{Cache}

The cache~\cite{burks1946preliminary} serves as a crucial component in computer science designed to temporarily store data, enhancing the efficiency of future request processing. Its main goal is to speed up access to frequently used data, thus minimizing dependence on slower storage disks. Popular caching policies include Least Recently Used (LRU)~\cite{mccabe1965serial} and First-In-First-Out (FIFO)~\cite{King71a}, with studies indicating LRU's superiority over FIFO~\cite{albers2002paging,dan1990approximate,chrobak1999lru}. The caching concept is widely applied in various fields, such as Databases~\cite{altinel2003cache} and CDN~\cite{harchol1999resource}.

\subsection{Warehouse Storage Strategy}

Automated Storage and Retrieval Systems (AS/RS) have attracted attention for their potential to enhance warehouse efficiency and reduce operational costs~\cite{GuGM07, RoodbergenV09}.
Various strategies for assigning items to storage locations have been widely adopted and evaluated~\cite{RoodbergenV09,AzadehKR19}. The random storage policy allocates each item type to a randomly chosen storage location and offers high space utilization~\cite{park2001optimal}. The closest storage policy places new items at the nearest available storage location to minimize the immediate travel distance~\cite{fukunari2008heuristic, gagliardi2012storage}. The turnover-based storage policy assigns items to storage locations based on their demand frequency~\cite{yuan2019velocity,li2020storage}.

\begin{figure}[t!]
\centering
\includegraphics[width=0.3\textwidth]{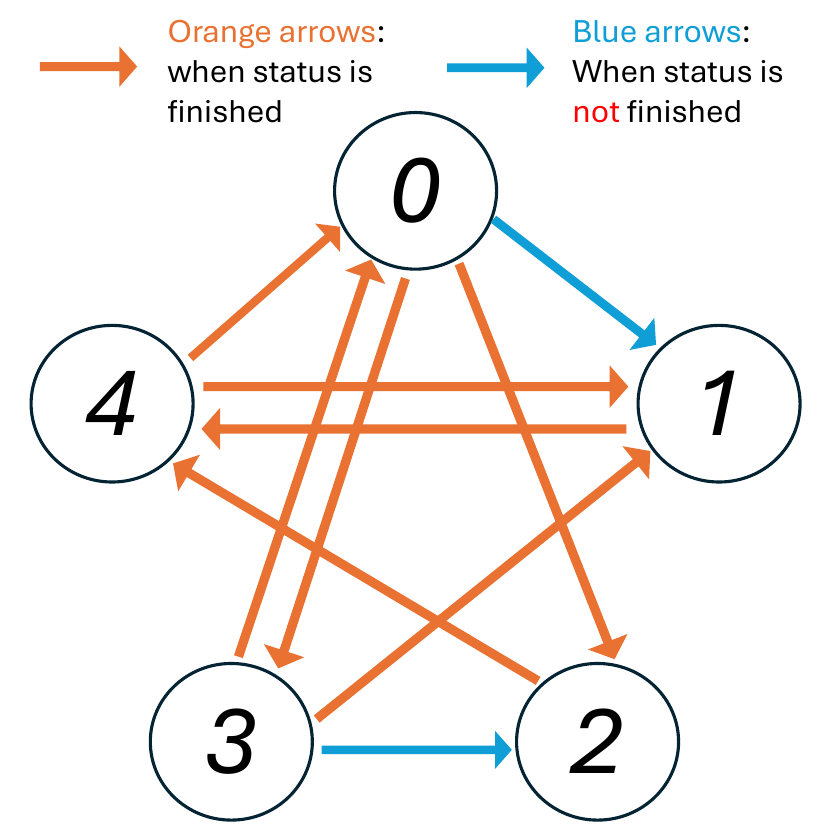} 
\caption{\textbf{Status 0}: If the agent's task item is not in the cache when assigned, it must retrieve the item from a shelf.
\textbf{Status 1}: If the agent's task item is in cache when assigned, it should retrieve the item from the cache.
\textbf{Status 2}: If at least one writable cache exists when the agent retrieves the item from the shelf, the agent needs to insert the item into cache.
\textbf{Status 3}: If no writable cache is available when an agent gets the item from the shelf, the agent goes to the unloading port.
\textbf{Status 4}: After retrieving or inserting the item from/to the cache, the agent heads to the unloading port.
}
\label{fig:state_machine}
\end{figure}

\section{Method}

In this section, we introduce our Caching-Augmented Lifelong MAPF (CAL-MAPF) framework along with a new map grid design, a TA and a cache lock mechanism. 

\subsection{Cache Grid}

As shown in \Cref{fig:cal_mapf}, we assume each shelf \(b_i \in B\) stores an infinite number of only one unique type of item. There will be a total of \(M\) different types of items corresponding to the number of shelves. Each unloading port \(u_k\) is associated with an independent task queue \(Q_k\). Each task \(q_i \in Q_k\) must be delivered to the unloading port \(U_k\). We will also involve a new type of map grid: caches. These additional vertices, designated as \( C=\{c_i | c_i \in V\} \) in purple, serve as interim storage areas to streamline the completion of tasks by reducing the travel time of the agents retrieving items. 

To the best of our knowledge, no previous MAPF works have explored this new cache mechanism, so we adopt the following simplifying assumptions:
(1) The warehouse has an unlimited supply of items, eliminating worries about stock running out.
(2) Agents have infinite capacity to carry items. However, they are limited to transporting one type of item at a time, enabling them to move any quantity of that specific item without restriction.
(3) Items removed or evicted from the cache are treated as if they vanish immediately.

\subsection{Task Assigner}
The task assigner (TA) is external to MAPF algorithm but can define all agent's target locations at any timestep. 
When the TA encounters a new task in $Q$ requiring completion, the first step typically determines where to get the item and which location should be allocated to an available agent. So, the TA first checks whether the task item exists in the cache grids. If so, TA will assign an agent with the cache location of the item. After the agent gets the item, TA then assigns the unloading port location to it. If not in cache grids, TA will give the item's shelf location to the agent and then one insertable cache grid location. Subsequently, the item is stored in the cache grid. Once stored, the agent can transport the item from the cache to an unloading port.  

However, similar to caching in computer architecture, there is a risk that other agents could replace the item in the cache with other items in multi-agent scenarios. This situation could result in MAPF algorithm, such as LaCAM, generating unusual collision-free plans at each timestep, causing agents to move back and forth between two adjacent grids. Thus, a lock mechanism ensures agents can secure an item after confirming its availability in cache grids.

\subsection{Cache Lock Mechanism}

In the CAL-MAPF framework, TA has a cache lock mechanism supported by a state machine to ensure efficient and synchronized access to cache locations. This approach effectively prevents race conditions during cache interactions. The mechanism uses read locks for agents fetching items and writes locks for agents updating the cache.
Every cache grid will have its independent read lock and write lock. This streamlined system guarantees that agents can access and modify cache contents safely, facilitating seamless task execution within the dynamic warehouse environment.

\subsubsection{Lock Types}
The cache lock mechanism is designed around two principal lock types, facilitating controlled access to the cache locations:
(1) \textbf{Read Lock (Shared Lock):} This type of lock permits an agent to access a cache location to retrieve items. Multiple agents can simultaneously hold a read lock on a single cache location as long as no write lock is active on that location. This arrangement ensures that when agents need to retrieve items from the cache, the item remains unchanged until all agents have successfully retrieved it.
(2) \textbf{Write Lock (Exclusive Lock):} An agent is required to secure a write lock prior to inserting an item into a cache location. Write locks are exclusive, implying that once an agent possesses a write lock on a cache location, no other agent can acquire either a read lock or another write lock on that location. This ensures that an agent can insert an item without impacting other agents.

\subsubsection{Lock Acquisition and Release}
The mechanism defines a protocol for lock acquisition and release:
(1) \textbf{Acquisition:} To obtain a read lock, an agent must verify no write lock is active on the desired cache location. Conversely, to secure a write lock, an agent must confirm that the target cache location is free from any active read or write locks.
(2) \textbf{Release:} Agents release read locks after successfully retrieving items. Agents release write locks once they have updated the cache, making the inserted new items available to others.

The cache lock mechanism ensures that no agent needs to return to shelf when its target location is in cache or unloading port, no two agents access the same cache block concurrently for writing, and multiple agents may read from the same cache block concurrently if no write lock is active. This concurrency control is critical to prevent race conditions and maintain the cache's consistency.

\begin{algorithm}[!t]
\small
\caption{CAL-MAPF with One Group Overview}
\label{alg:cache_algo}
\textbf{Input}: Map \(G\), Initial Locations \(S = s_i\), Task Queue \(Q\), Agents \(A\)
\begin{algorithmic}[1]
\State portLoc = getPortLoc(G)
\For{agent in agents}
    \State agent.startLoc = $s_i$
    \State agent.task = Q.pop()
    \State agent.targetLoc = getMapLoc(agent.task)
    \State agent.status = 0
\EndFor
\While{not \(Q\).empty()}
    \State plan = MAPF(agents)
    \State agents.excute(plan)
    \State AgentReleaseLocks(agents, cache)
    \State AgentGetLocks(agents, cache)
\EndWhile
\end{algorithmic}
\end{algorithm}

\begin{algorithm}[!t]
\small
\caption{Task Assigner Functions}
\label{alg:cache_algo_func}
\begin{algorithmic}[1]
\Function{AgentReleaseLocks}{agents, cache}
    \For{agent in agents}
        \State agent.startLoc = plan.endLoc[agent]
        \If{agent.targetLoc == agent.startLoc}
            \If{agent.status == 1 or 2}
                \State agent.status = 4
                \State cache.releaseAllLock(agent)
                \State agent.targetLoc = portLoc
            \EndIf
        \EndIf
    \EndFor
\EndFunction

\Function{AgentGetLocks}{agents, cache}
\For{agent in agents}
    \If{agent.status == 0}
        \If{agent.targetLoc == agent.startLoc}
            \State targetLoc = cache.insert(agent)
            \State CheckTarget(agent, targetLoc, 3, 2)
        \Else
            \State targetLoc = cache.check(agent)
            \State CheckTarget(agent, targetLoc, 0, 1)
        \EndIf
        \State continue
    \EndIf

    \If{agent.status == 3}
        \If{agent.targetLoc == agent.startLoc}
            \State AgentReachPort(agent)
        \Else
            \State targetLoc = cache.insert(agent)
            \State CheckTarget(agent, targetLoc, 3, 2)
        \EndIf
        \State continue
    \EndIf

    \If{agent.status == 4}
        \If{agent.targetLoc == agent.startLoc}
            \State AgentReachPort(agent)
        \EndIf
    \EndIf
\EndFor
\EndFunction

\Function{AgentReachPort}{agent}
\State agent.task = Q.pop()
\State targetLoc = cache.check(agent)
\State CheckTarget(agent, targetLoc, 0, 1)
\EndFunction

\Function{CheckTarget}{agent, targetLoc, statusa, statusb}
\State agent.status = statusa
\If{targetLoc != portLoc}
    \State agent.status = statusb
\EndIf
\State agent.targetLoc=targetLoc
\EndFunction

\end{algorithmic}
\end{algorithm}

\begin{algorithm}[!h]
\small
\caption{Cache Operation Functions}
\label{alg:cache_algo2}
\begin{algorithmic}[1]
\Function{releaseAllLock}{agent}
    \State cacheGrid = findCache(agent.startLoc)
    \State cacheGrid.item = agent.task
    \State cacheGrid.readLock.remove(agent)
    \State cacheGrid.writeLock.remove(agent)
    \State incoming.remove(taskItemId)
\EndFunction
\Function{Check}{agent}
\For{cacheGrid in cache.allGrids}
    \State noWrite = cacheGrid.writeLock.empty()
    \State hasItem = cacheGrid.item == agent.task
    \If{noWrite and hasItem}
        \State cacheGrid.readLock.add(agent.id)
        \State \Return cacheGrid.loc
    \EndIf
    \State \Return portLoc
\EndFor
\EndFunction
\Function{Insert}{agent}
    \State isInCache = cache.findItem(agent)
    \State isIncoming = incoming.find(agent.task)
    \If{isInCache or isIncoming}
        \State \Return portLoc
    \EndIf
    
    \State cacheGrid = cache.findEmptyCache()
    \If{cacheGrid != -1}
        \State cacheGrid.writeLock.add(agent.id)
        \State \Return cacheGrid.loc
    \EndIf

    \State candidates = sortByCacheEvitedPolicy(cache.allGrids)
    \For{cacheGrid in candidates}
        \If{cacheGrid.readLock.empty()}
            \If{cacheGrid.writeLock.empty()}
                \State incoming.append(agent.task)
                \State cacheGrid.writeLock.add(agent.id)
                \State \Return cacheGrid.loc
            \EndIf
        \EndIf
    \EndFor
    \State \Return portLoc
\EndFunction
\end{algorithmic}
\end{algorithm}

\subsection{Algorithm}

As shown in \Cref{alg:cache_algo,alg:cache_algo_func,alg:cache_algo2}, we will introduce the whole procedure of CAL-MAPF. It is mainly based on the TA, who encounters new tasks and decides what locations the agents need to visit. As illustrated in \Cref{fig:cal_mapf}, CAL-MAPF could have multiple unloading ports and we call each unloading port and cache around the port a group. Each group is independent in task queue, cache grids, and agents. Agents in one group can only accept tasks from this group and use cache grids of this group. \Cref{alg:cache_algo} illustrates the operation of one group. 

\begin{figure*}[htb]
\centering
\includegraphics[width=\textwidth]{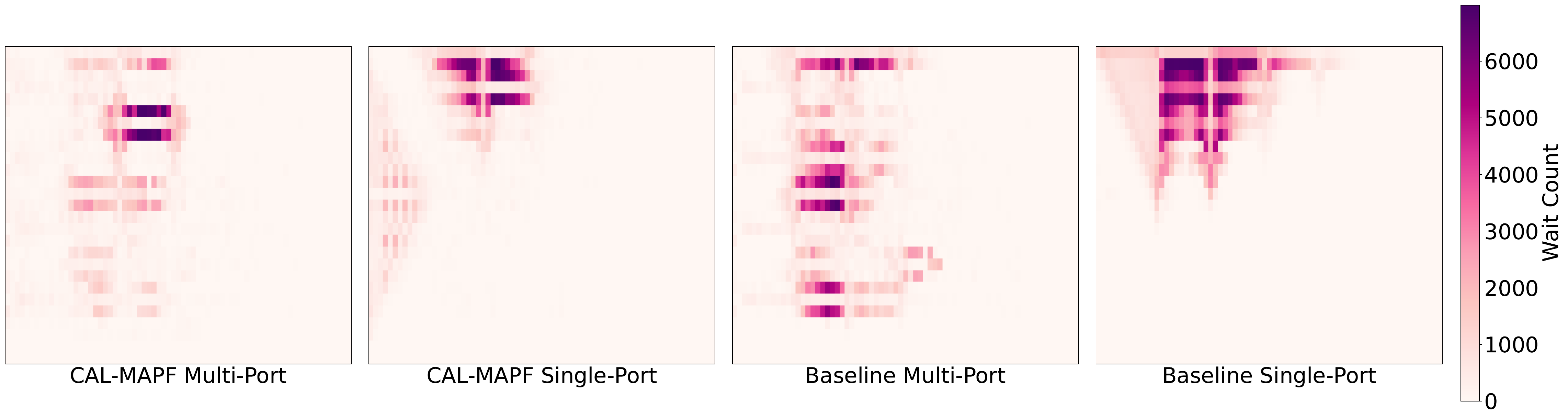} 
\caption{The average frequency of agent wait actions on each map grid with 256 agents under Zhang distribution. As the agent number is large enough, both CAL-MAPF and baseline experience severe traffic congestion. The congestion position bias in the map could be caused by the randomization of item indexes and tasks in $Q$.}
\label{fig:WaitMaps}
\end{figure*}

TA uses a state machine to control agent status and how they transfer to another based on different situations. When any agent reaches its target location, TA must check all agents, update their status, and assign new target locations. The state machine is illustrated in \Cref{fig:state_machine}, and the pseudocode of \Cref{alg:cache_algo} outlines how an agent transitions through five states while performing the MAPF algorithm for a group. The TA monitors and updates the status of agents based on the item's location, the state of the cache, and the agents' current status. This algorithm ensures the efficient completion of all tasks by orchestrating the agents' paths and their interactions with the cache and unloading port.

Overall, CAL-MAPF needs to use TA to assign target locations to all spare agents based on $Q$. Then invoke MAPF solver to find a collision-free solution and execute it until at least one agent reaches its target location. Then TA will check all agent's statuses, release locks and change agent status following the state machine. When changing status, TA needs to use tasks in $Q$ and cache replacement policy to determine target locations for agents and acquire locks for agents. CAL-MAPF repeats this process during the lifelong scenario.

In \Cref{alg:cache_algo}, we first set the status of each agent to 0 and assign them initial target locations (Lines 1-6). Next, we obtain collision-free paths for each agent by invoking a MAPF algorithm. The plan will be executed until at least one agent reaches its target location (Lines 7-9). Upon arrival, TA updates the lock information, attempts to assign new target locations to available agents, and updates the status of all agents based on the current warehouse situation (Lines 10-11).

TA should first attempt to release locks to allow other agents the opportunity to access cache grids. Status 1 is trying to read a cache grid, and Status 2 is going to write. Only agents in status 1 and status 2 can hold locks. After reaching their target location, they must go to the unloading port to deliver items. So, we verify whether these agents have reached their targets before releasing their held locks (\Cref{alg:cache_algo_func} Lines 1-8). 

Then we evaluate whether other agents can alter their status. For agents in status 0, heading to shelves, we examine if they have reached the shelves and gotten items. We then check if they can do a write operation in the cache and apply a cache replacement policy to select a cache grid. Thus, status 0 may transition to status 2 (write to cache) or Status 3 (proceed directly to unloading port) (\Cref{alg:cache_algo_func} lines 9-14). If they have not yet arrived, we check whether they can retrieve an item from the cache, which would change their status to 1 (read cache) (\Cref{alg:cache_algo_func} Lines 15-18). We do the same status changes for agents in Status 3 and Status 4 as we describe in ~\Cref{fig:state_machine} (\Cref{alg:cache_algo_func} Lines 19-28).
\begin{figure*}[htb]
\centering
\includegraphics[width=\textwidth]{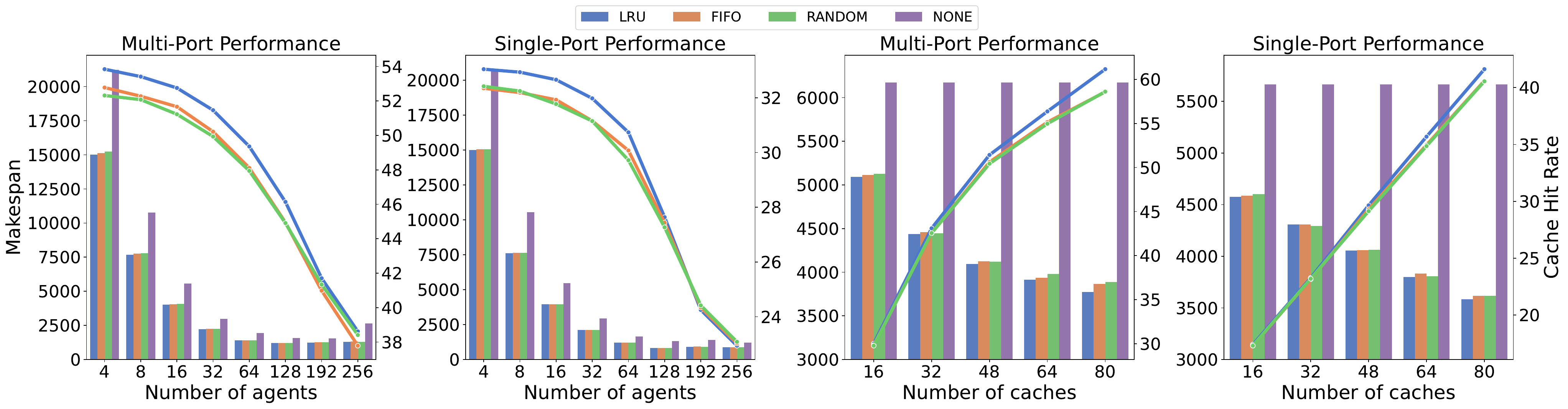} 
\caption{Makespan (Bar chart, lower is better) and Cache Hit Rate (Line chart, higher is better). LRU, FIFO, and RANDOM represent CAL-MAPF with different cache replacement policies. NONE represents Lifelong MAPF without cache. 
}
\label{fig:nagent_ncache_makespan_all}
\end{figure*}

In \Cref{alg:cache_algo2}, we detail the process of reading or insertion (writing) a new item into cache grids using a cache replacement policy. For the read operation, it is necessary to verify if the item exists in the caches and if there is an existing write lock on the cache grid (\Cref{alg:cache_algo2} lines 7-14). For the write operation, we first check whether the item is already in the cache area or if another agent plans to insert the same item into the caches (\Cref{alg:cache_algo2} lines 15-19). Then, we determine if an available cache exists for inserting the item (\Cref{alg:cache_algo2} Lines 20-31).



\section{Experimental Results}

We use Lifelong MAPF without cache as a baseline, which aligns with our problem definition. To evaluate performance, we compare CAL-MAPF with Lifelong MAPF, both using LaCAM as the MAPF solver for generating collision-free paths. CAL-MAPF and Lifelong MAPF were implemented in C++, building on parts of the existing LaCAM codebase\footnote{The LaCAM source code is publicly accessible at \url{https://github.com/Kei18/lacam}. Our implementation is available at \url{https://github.com/HarukiMoriarty/CAL-MAPF}.}
All experiments were conducted on a system running Ubuntu 20.04.1, equipped with an AMD Ryzen 3990X 64-core CPU and 64GB RAM at 2133 MHz.

\subsection{Test Settings} 

As shown in \Cref{fig:cal_mapf}, we designed a warehouse map (27x71) with caches based on our problem definition. The map is adapted from the warehouse map of MAPF benchmark~\cite{stern2019multi}. It includes 1600 shelf grids, a maximum of 80 cache grids and 4 unloading ports. The maximum cache-to-shelf ratio is 5\%. We tested CAL-MAPF in both multi-port and single-port scenarios, as depicted in \Cref{fig:cal_mapf}. The multi-port scenario has 4 working groups of unloading ports and cache grids, each with a maximum of 20 cache grids. The single-port scenario maintains the same number of total cache grids and agents as the multi-port but only has one unloading port. Because each shelf grid represents a unique kind of item, and we randomly assign an index to each shelf grid. We test all scenarios with different cache numbers \{16, 32, 48, 64, 80\} by deleting some cache grids (described in \Cref{fig:cal_mapf}). We have also chosen various total numbers of agents, \(\{4, 8, 16, 32, 64, 128, 192, 256\}\). In the multi-port scenario, each group has an equal number of agents \(\{1, 2, 4, 8, 16, 32, 48, 64\}\).

Since we can expect the distribution of the task queue could significantly affect the performance of cache design, we designed three input task distributions to test CAL-MAPF, including: 
(1) $M$-$K$ distribution (MK): For any consecutive subarray of length $M$ in the task queue $Q$, there are at most $K$ different kinds of items. This distribution is inspired by~\cite{albers2002paging}, where LRU has been proven to have an upper bound on the cache miss rate and to be better than FIFO. This distribution can also represent several items people purchase daily, and some items may become very popular at one time, replacing previously popular items. 
(2) 7:2:1 distribution (Zhang): There are 70\% kinds of items with only a 10\% appearance probability in the task queue, 20\% kinds of items with a 20\% probability, and 10\% with a 70\% appearance probability~\cite{zhang2016correlated}. 
(3) Real Data Distribution (RDD): We obtain data from Kaggle's public warehouse data\footnote{kaggle.com/datasets/felixzhao/productdemandforecasting}, build a probability distribution based on the frequency of data and generate tasks from this probability distribution. 

\begin{figure*}[htb]
\centering
\includegraphics[width=\textwidth]{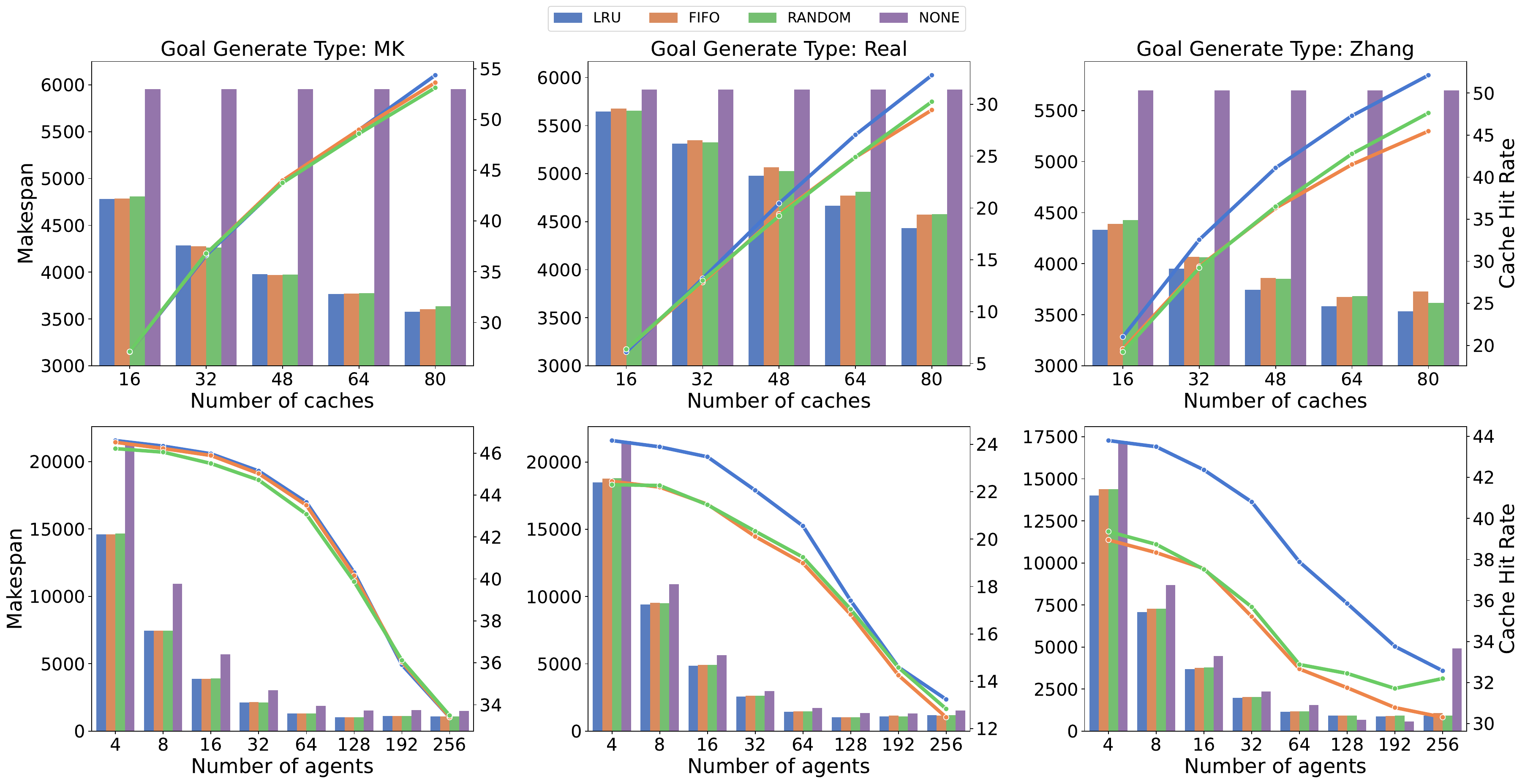} 
\caption{Makespan (Bar chart, lower is better) and Cache Hit Rate (Line chart, higher is better). As the number of caches increases, CAL-MAPF's cache hit rate and makespan performance improve. In the MK, Real, and Zhang distributions, CAL-MAPF surpasses the baseline in most test settings. However, it is observable that as the number of agents increases, the improvement offered by CAL-MAPF diminishes. Additionally, an abnormal increase in makespan is observed at baseline under the Zhang distribution with 256 agents.
}
\label{fig:nagent_makespan_all_goal}
\end{figure*}

\begin{figure}[t!]
\centering
\includegraphics[width=0.4\textwidth]{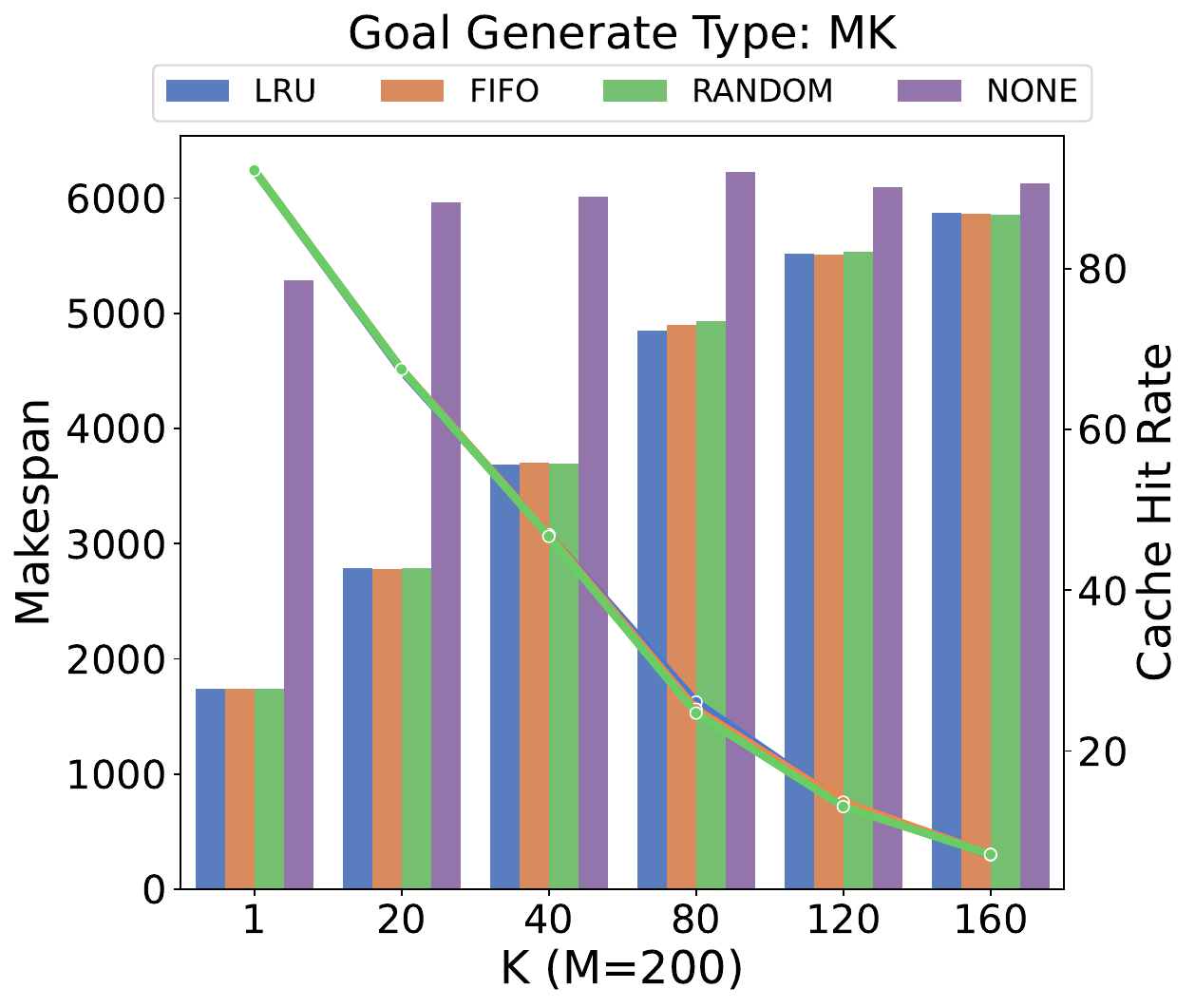} 
\caption{Makespan (Bar chart, lower is better) and Cache Hit Rate (Line chart, higher is better). The hit rate significantly depends on the input task distribution, and the hit rate also greatly affects the final makespan performance.
}
\label{fig:nk_makespan_MK}
\end{figure}

We randomly generate agent start locations and \(Q\) for all input task distributions with a total length of 1,000. For the MK distribution, we select \(M=200\) and \(K\) values of \(\{1,20,40,80,120,160\}\). We allocate a total of 10 seconds for the MAPF solver to find a collision-free solution for all 1,000 tasks in \(Q\). We use makespan, the completion time when the last task has been accomplished, to show performance. For CAL-MAPF, we have tested three different cache replacement policies. Least Recently Used (LRU), First-In-First-Out (FIFO), and RANDOM. We use NONE to represent the baseline method.

\subsection{Performance} 


Many variables can influence the performance of CAL-MAPF, including the number of agents, the number of caches, and the distributions of input tasks. As illustrated in \Cref{fig:nagent_ncache_makespan_all}, increasing the number of caches improves cache hit rate and makespan performance for both single- and multi-port scenarios. And the cache hit rate decreases as the number of agents grows. It is also observed that as the number of agents continues to increase (from 64 to 256), there is not much difference in the makespan for both CAL-MAPF and the baseline, and the cache hit rate is dropping.


Two reasons could contribute to this scenario: (1) CAL-MAPF and baseline's makespan remain at the same level or even worse with too many agents: traffic congestions occur in the map as the number of agents increases. 
(2) The CAL-MAPF cache hit rate continues to decrease as the makespan performance shows minimal differences: We employ a lock mechanism to maintain items stored in the cache area. Traffic congestion prevents many agents from reaching their target in the cache. For other agents not significantly affected by traffic congestion, they are constantly assigned new tasks. They cannot obtain locks in the cache because locks are held by agents in congestion. So the item distribution cannot update with input tasks, which ultimately leads to a continuous drop in the cache hit rate.
This second reason can also clarify why CAL-MAPF performs better as the number of caches increases: enlarging the cache area mitigates traffic congestion and enhances the cache hit rate.

In \Cref{fig:nagent_makespan_all_goal} and \Cref{fig:nk_makespan_MK}, we present the performance of CAL-MAPF and the baseline for various task distributions. \Cref{fig:nagent_makespan_all_goal} demonstrates that as the number of caches increases, CAL-MAPF's performance improves. 
In the MK, Real, and Zhang distributions, CAL-MAPF surpasses the baseline in most test settings. However, it is observable that as the number of agents increases, the improvement offered by CAL-MAPF diminishes.
One possible reason is that the low hit rate leads agents to require longer paths to complete a task. As depicted in \Cref{fig:nk_makespan_MK}, the hit rate significantly depends on the input task distribution, and the hit rate also greatly affects the final makespan performance. 

Furthermore, in \Cref{fig:nagent_makespan_all_goal}, an abnormal increase in makespan is observed at baseline under the Zhang distribution with 256 agents. Traffic congestion could contribute to this abnormally poor performance. \Cref{fig:WaitMaps} illustrates the average frequency of agent wait actions on each grid when the number of agents reaches 256 with the Zhang distribution. Both CAL-MAPF and baseline experience severe traffic congestion. 

High cache hit rates and smooth traffic are crucial for the performance of CAL-MAPF. We can increase the number of caches to enhance the cache hit rate. And we can also directly add more agents to improve the makespan performance. However, both methods come with their drawbacks. The number of caches is constrained by the space available close to the unloading port. Introducing more agents can lead to severe traffic congestion, ultimately causing the hit rate to drop.
Nevertheless, there are potential solutions to mitigate these adverse effects, such as implementing more efficient map designs~\cite{ijcai2023p611} and using one-way systems near the cache and unloading ports~\cite{zhang2024guidance}. Additionally, as we currently employ a simple task assignment policy, we could implement a more advanced policy with predictive capabilities, leveraging real warehouse task data to enhance the cache hit rate. The cache replacement policies, such as LRU and FIFO, may be overly simplistic for CAL-MAPF. Incorporating more complex policies, including learning-based approaches, could further improve the cache hit rate, especially since CAL-MAPF allows more planning time than traditional caches in computer architecture.

\section{Conclusion}

This work presents a novel mechanism, Caching-Augmented Lifelong MAPF (CAL-MAPF), designed to improve the performance of Lifelong MAPF. We have developed a new map grid type called cache for temporary item storage and replacement. Additionally, we have devised a locking mechanism for caches to improve the stability of the planning solution. This cache mechanism was evaluated using various cache replacement policies and a spectrum of input task distributions. We identified three main factors that significantly impact CAL-MAPF performance through experimentation: suitable input task distribution, high cache hit rate, and smooth traffic. In general, CAL-MAPF has demonstrated potential for performance improvements in certain task distributions and map and agent configurations.


\bibliographystyle{IEEEtran} 
\bibliography{strings,myref}

\end{document}